\documentclass[journal]{IEEEtran}
\ifCLASSINFOpdf
\else
\fi

\usepackage{graphics} 
\usepackage{epsfig} 
\usepackage{mathptmx} 
\usepackage{times} 
\usepackage{amsmath} 
\usepackage{amssymb}  
\usepackage{makecell}
\usepackage{booktabs}
\usepackage{cite}
\usepackage{diagbox}
\usepackage{bbding}
\usepackage{enumitem}

\begin{document}
%
\title{A Bi-directional Adaptive Framework for Agile UAV Landing}
%
%
%

 \author{Chunhui Zhao$^{\dagger}$,~\IEEEmembership{Member,~IEEE,},~Xirui Kao$^{\dagger}$,~Yilin Lu,~Yang Lyu$^{*}$,~\IEEEmembership{Member,~IEEE}
 \thanks{$^{\dagger}$C. Zhao and X. Kao contributed equally to this work.}%
 \thanks{
 Chunhui Zhao, Xirui Kao, Yilin Lu, and Yang Lyu are with the
 School of Automation, Northwestern Polytechnical University, Xi’an Shanxi,
 710072 China. email: lyu.yang@nwpu.edu.cn.
 This work was supported by the National Natural
 Science Foundation of China under Grant 62203358, 62233014, and 62073264.
 }
 }

%
%

\markboth{Journal of \LaTeX\ Class Files,~Vol.~14, No.~8, August~2015}%
{Shell \MakeLowercase{\textit{et al.}}: Bare Demo of IEEEtran.cls for IEEE Journals}
%



\maketitle

\begin{abstract}
Autonomous landing on mobile platforms is crucial for extending quadcopter operational flexibility, yet conventional methods are often too inefficient for highly dynamic scenarios. The core limitation lies in the prevalent ``track-then-descend'' paradigm, which treats the platform as a passive target and forces the quadcopter to perform complex, sequential maneuvers. This paper challenges that paradigm by introducing a bi-directional cooperative landing framework that redefines the roles of the vehicle and the platform.
The essential innovation is transforming the problem from a single-agent tracking challenge into a coupled system optimization. Our key insight is that the mobile platform is not merely a target, but an active agent in the landing process. It proactively tilts its surface to create an optimal, stable terminal attitude for the approaching quadcopter. This active cooperation fundamentally breaks the sequential model by parallelizing the alignment and descent phases. Concurrently, the quadcopter's planning pipeline focuses on generating a time-optimal and dynamically feasible trajectory that minimizes energy consumption. This bi-directional coordination allows the system to execute the recovery in an agile manner, characterized by aggressive trajectory tracking and rapid state synchronization within transient windows. The framework's effectiveness, validated in dynamic scenarios, significantly improves the efficiency, precision, and robustness of autonomous quadrotor recovery in complex and time-constrained missions.
 \end{abstract}


%
\IEEEpeerreviewmaketitle

\section{Introduction}

With the rapid development of advanced perception and control technologies, autonomous quadcopters have shown great application potential in many fields, such as emergency search and rescue, logistics distribution, and high-altitude inspection, due to their excellent maneuverability, hovering ability, and ease of deployment\cite{r1,r2,r3}. Despite these advantages, there is a common endurance bottleneck. Typically, the flight time of a quadrotor during one battery cycle is only 15 to 25 minutes, an inherent limitation that greatly restricts its operational range and mission continuity, becoming a primary obstacle to its large-scale, long-duration application in the real world.
An effective solution to break through this limitation is to enable autonomous recharging or recycling of the quadcopter's battery on mobile platforms, such as an Autonomous Ground Vehicle (AGV) or Autonomous Surface Vehicle (ASV). The cooperation between the heterogeneous platforms can unleash unprecedentedly large-scale and highly efficient operations in more complex environments \cite{wang2023cooperative,wu2020cooperative}. 
\begin{figure}[t]
\vspace*{2mm} 
\centering
\includegraphics[width=\columnwidth]{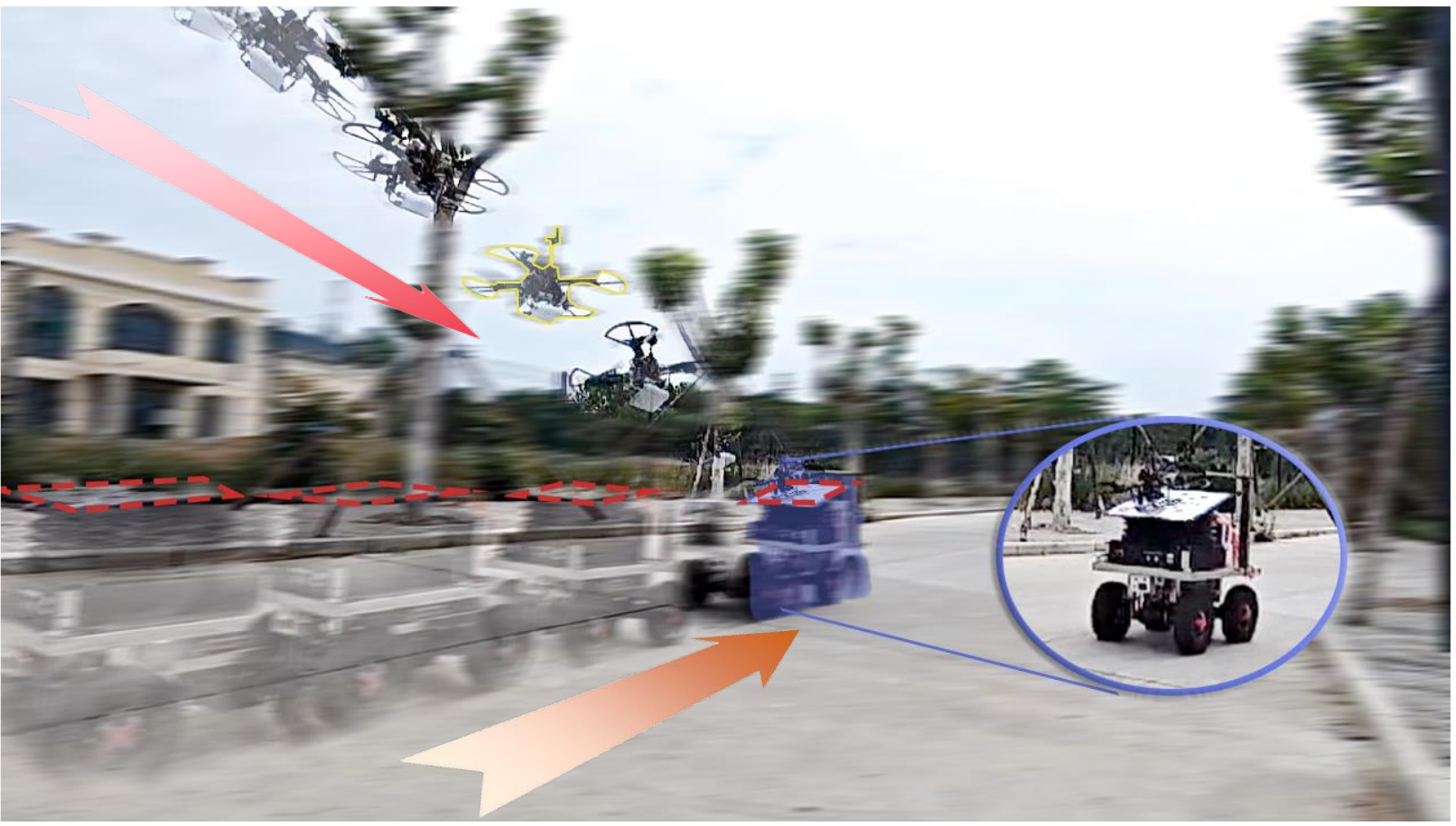}
\caption{Cooperative landing process on a mobile variable-attitude platform.}
\label{fig:exp_hardware}
\end{figure}

Achieving reliable autonomous landing on mobile platforms is not merely a technical requisite for cooperation; it is a critical bottleneck fraught with formidable challenges. The core difficulty lies in the platform's unpredictable dynamics. Buffeted by environmental forces—from rough terrain for AGVs to volatile waves for ASVs—the platform's motion injects severe uncertainty into the landing task \cite{r4,r5}. This volatility culminates in the most daunting scenario for time-critical missions: the emergence of fleeting, transient ``landing windows''--- moments of relative stability that represent the only opportunity for a safe maneuver. Missing this window often means mission failure. Therefore, the central challenge this work addresses is not merely improving general performance, but engineering the requisite dynamic responsiveness and efficiency to reliably seize these ephemeral opportunities.

\begin{figure}[t]
\vspace*{2mm} 
\centering
\includegraphics[width=\columnwidth]{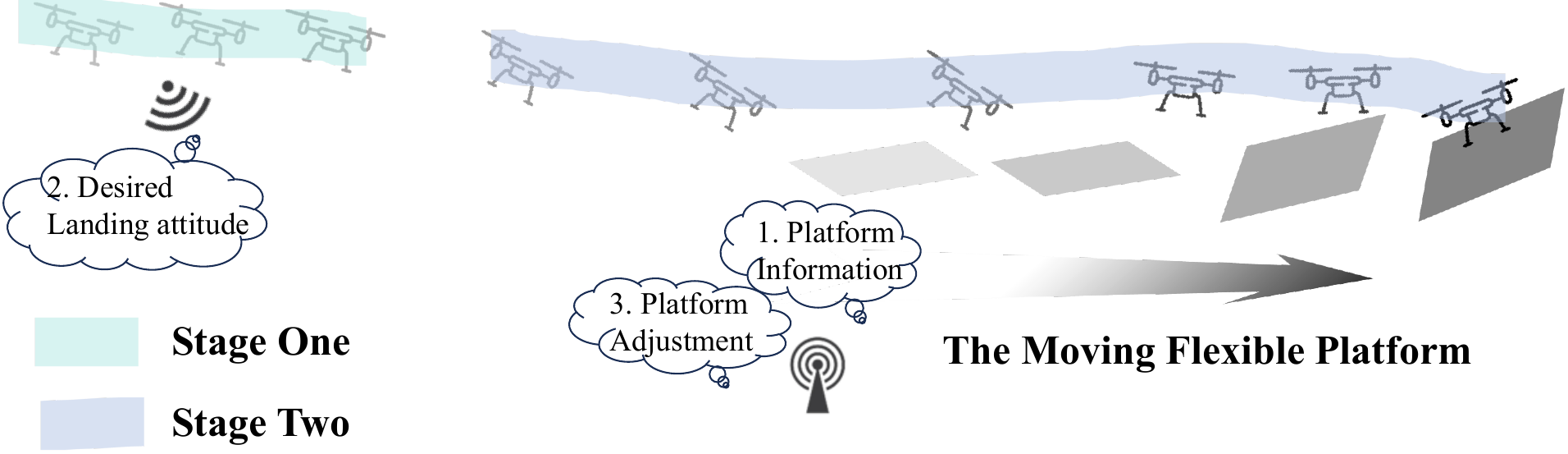}
\caption{The process occurs in two stages. First, in Stage One, the quadcopter obtains information from the platform to calculate its landing attitude and synchronizes this data back. Then, in Stage Two, the quadcopter plans its trajectory and begins to land as the platform simultaneously adjusts its own attitude to facilitate the landing.}
\label{fig:TwoStage}
\end{figure}

This paper proposes a bi-directional cooperative landing strategy that integrates trajectory optimization with active platform control to minimize recovery time. By leveraging the platform's controllable attitude to reshape the feasible trajectory space, the framework enables an agile landing manner. In this context, agility is defined by the system's dynamic responsiveness and its ability to seize transient landing windows through time-optimal maneuvers, rather than absolute ground velocity. This cooperation facilitates an aggressive ``sprint-then-brake'' pattern, replacing conventional cautious descents where relative velocity nears zero.To manage the control complexity of this coupled system, we decompose the process into two synergistic stages (Fig. \ref{fig:TwoStage}). Stage one involves the collaborative planning of optimal terminal states, including the crucial platform tilt angle and the drone's final attitude. With these targets defined, stage two focuses on optimizing a time- and energy-efficient trajectory for the quadrotor to achieve them.

The contributions of this paper can be summarized as follows: \begin{itemize} 
\item A novel cooperative landing framework is proposed, which treats the quadrotor and the mobile platform as a coupled system with common terminal states, especially an adaptive terminal attitude. 
\item A two-stage cooperative planning algorithm is designed to online generate time-optimal, dynamically feasible, large-maneuver trajectories while optimizing for landing stability and safety. 
\item The proposed framework is comprehensively validated through high-fidelity simulations and physical experiments, which demonstrate its advantages over classic baseline methods. 
\end{itemize}


\section{Related Works}
\subsection{Vision-based Relative State Estimation and Landing}
Autonomous landing on moving platforms presents a significant challenge due to the underactuated and nonholonomic nature of quadrotors, where attitude dynamics are tightly coupled with translational acceleration. Early research primarily focused on unilateral, passive adaptation, where the quadrotor adjusts itself to a non-cooperative platform. 
A prominent category utilizes visual servoing \cite{r6,r7,r8}. These methods leverage real-time image errors for feedback control, offering advantages such as rapid response rates and low computational overhead. 
However, their reactive nature implies that the generated paths are rarely time-optimal. Furthermore, these approaches rely heavily on continuous visibility of target features to maintain stability. This dependency creates a critical bottleneck when addressing the stringent timeliness requirements of the ``landing window'' considered in this work, where feature loss or aggressive maneuvering can lead to mission failure.
\subsection{Trajectory Planning for Dynamic Landing}
To overcome the myopia of reactive control, another category focuses on trajectory planning \cite{r9,r10,r11}. By incorporating state estimation and motion prediction, these methods can generate feasible paths for landing even on high-speed vehicles. Pushing the limits of maneuverability, Kaufmann et al. \cite{r19} and Habas et al. \cite{r20} achieved aggressive maneuvers, such as flips, using onboard sensors driven by deep learning policies. 
Despite these achievements, deep learning-based approaches often demand substantial computational resources, which challenges the payload constraints of small-scale robots, and lack theoretical reliability guarantees. Moreover, in standard trajectory planning frameworks, the platform is typically treated as a passive target. The objective is restricted to reaching a moving coordinate, ignoring the potential for interactive coordination. This lack of active attitude synchronization limits the system's ability to execute time-critical maneuvers safely.
\subsection{Cooperative Control of Heterogeneous Systems}
Recognizing the limitations of unilateral approaches, recent research has pivoted towards cooperative strategies.
One direction involves hardware-based cooperation, specifically the design of specialized mechanical landing gear \cite{r1,r2,r3}. While customized mechanisms can effectively mitigate landing impacts, they incur a significant weight penalty—--ranging from 10\% \cite{r21} to 40\% \cite{r22} of the aircraft's mass—which inevitably compromises flight endurance.

Alternatively, algorithmic cooperation seeks to coordinate the motion of heterogeneous agents. Meng et al. \cite{r12} fused multi-sensor data for the shipboard landing of fixed-wing vehicles. While fixed-wing platforms offer superior endurance, they lack the hover capability and maneuverability of quadrotors, making precise recovery more complex. Rabelo et al. \cite{r13} formulated the landing problem as a formation control task, requiring the quadrotor to hover above the platform before descending. While stable, this ``hover-then-descend'' strategy is inherently time-consuming. In \cite{r23}, a coordination technique for a skid-steered AGV and a quadrotor was proposed, but its real-world feasibility remains unverified. Si et al. \cite{r14} designed a tracking and landing system, yet it lacked a reverse information flow from the quadrotor to the platform, precluding a fully integrated, bi-directional planning approach.

In summary, existing works are either limited to unilateral adaptation struggling with the tight constraints of short landing windows. or lack the dynamic responsiveness required for efficient cooperation. To bridge this gap, this paper proposes a cooperative framework based on a bi-directional adaptive strategy. By treating the platform as an active agent capable of attitude alignment, we integrate the descent and alignment processes, enabling the quadrotor to execute rapid, large-maneuver trajectories that are infeasible for traditional methods.

\begin{figure*}[htbp]
  \centering
  \includegraphics[width=\textwidth]{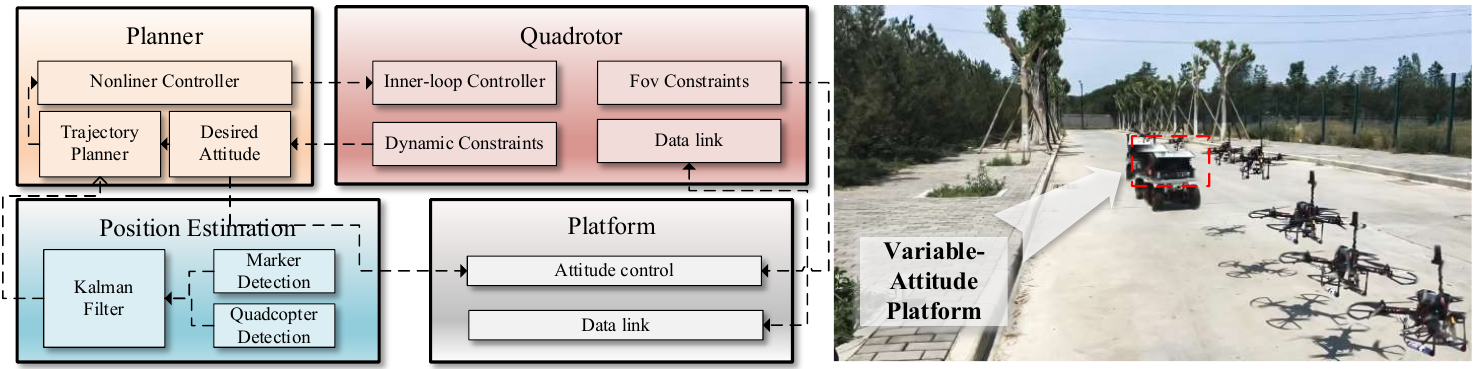}  
  \caption{The framework diagram of the coordinated landing system of the quadcopter and the landing platform. The quadrotor calculates the desired landing attitude based on its own constraints, and the platform can actively adjust the angle to adapt to the quadcopter's attitude during landing.}
  \label{fig:wide_figure}
\end{figure*}

\section{Trajectory Planning And System Control} 
\subsection{Two Stage Trajectory Planning}
In our cooperative landing scenario, we assume the landing stage can adaptively adjust its attitude and can dynamically change its inclination to coordinate the landing of the quadrotor. This feature introduces a new optimization dimension for trajectory planning, and the quadrotor no longer needs to be forced to reach a fixed platform attitude in the end, but can actively select an optimal terminal attitude to improve landing performance.

Focusing on the cooperative mechanism, we first propose a terminal attitude heuristic selection method based on dynamic optimization(as depicted in Fig. 2). The core idea of this method is to determine the optimal state of the quadrotor after completing the main translation maneuver phase by solving an optimization problem, thereby indirectly obtaining the ideal landing attitude.

In order to determine the terminal attitude, we established an optimization model designed to reflect the comprehensive performance of the quadrotor at the end of its main translation maneuver phase, and simplified the model into a two-dimensional model:

\begin{equation} \label{eq:lagrangian_cost_split}
	\begin{split}
		L(\mathbf{X}, \mathbf{U}) ={}& w_{a}\left[(a_{bx}(t))^2 + (a_{bz}(t))^2\right] \\
		& + w_j\left[j_{bx}(t)^2 + j_{bz}(t)^2\right]
	\end{split}
\end{equation}

Based on the above requirements, the optimization problem is formulated as follows:

\begin{subequations} \label{eq:full_optimization_problem}
	\begin{align}
		\min_{\mathbf{U}(t), T_f} \quad & J = \int_{t_0}^{T_f} L(\mathbf{X}(t), \mathbf{U}(t)) \,\mathrm{d}t + w_T \cdot T_f \label{eq:obj_func} \\
		\textrm{s.t.} \quad & \mathbf{\dot{X}}(t) = f(\mathbf{X}(t), \mathbf{U}(t)), \quad & \forall t \in [t_0, T_f] \label{eq:constraint_dynamics} \\
		& \mathbf{X}(t_0) = \mathbf{X}_0 \label{eq:constraint_initial} \\
		& p_{bx}(T_f) = x_f, \quad p_{bz}(T_f) = z_f \\
		& v_{bx}(T_f) = v_{xf}, \quad v_{bz}(T_f) = v_{zf} \\
		& v_{bx}(t)^2 + v_{bz}(t)^2 \le {v_{max}^2}, \quad & \forall t \in [t_0, T_f] \label{eq:constraint_vel_limit} \\
		& a_{bx}(t)^2 + a_{bz}(t)^2 \le {a_{max}^2}, \quad & \forall t \in [t_0, T_f] \label{eq:constraint_accel_limit} \\[0.5em] 
		& \left| \frac{a_{bx}(T_f)}{a_{bz}(T_f) + g} \right| \le \tan(\phi_{max}) \label{eq:constraint_platform_angle} \\[0.5em] 
		& T_f \ge \frac{|\phi_{opt}|}{\omega_{plat}} + \tau_{delay} \label{eq:constraint_actuation_time}
	\end{align}
\end{subequations}
where Eq.(\ref{eq:constraint_dynamics}) is the system dynamics constraint, $\mathbf{X}(t)=[p_{bx},p_{bz},v_{bx},v_{bz},a_{bx},a_{bz}]^T$, and $\mathbf{U}(t)=[j_{bx},j_{bz}]^T$ are the state and control vectors, respectively. The boundary constraints include the initial state Eq.(\ref{eq:constraint_initial}) and the terminal constraints Eq.(2d)-(2g). 

Crucially, to ensure the planned maneuver is physically feasible for the mobile platform, we impose two additional terminal constraints. Eq.(\ref{eq:constraint_platform_angle}) restricts the required terminal inclination $\phi_{opt}$  within the platform's mechanical tilt limit $\phi_{max}$. Furthermore, Eq.(\ref{eq:constraint_actuation_time}) enforces a lower bound on the maneuver duration $T_f$ based on the platform's actuation speed. Here, $\omega_{plat}$ denotes the maximum rotational velocity of the platform deck, and $\tau_{delay}$ accounts for system latency. This constraint guarantees that the platform has sufficient time to reach the target attitude before the quadrotor touches down, preventing infeasible commands where the required tilt rate exceeds the platform's capabilities.

With the optimal terminal acceleration $\mathbf{a}_{opt} = [a_{bx}(T_f), a_{bz}(T_f)]^T$ and the corresponding duration $T_f$ obtained from the optimization model in Eq. (\ref{eq:full_optimization_problem}), we proceed to the second stage: generating the full 3D continuous trajectory.

Following the computationally efficient motion primitive generation method proposed in \cite{r15}, we formulate the trajectory planning as a minimum jerk trajectory problem decoupled in three orthogonal axes. For each axis $k \in \{x, y, z\}$, the goal is to find a trajectory $p_k(t)$ that minimizes the jerk cost function:

\begin{equation} \label{eq:min_jerk_cost}
    J_{traj} = \sum_{k \in \{x,y,z\}} \int_{0}^{T_f} \left( \frac{d^3 p_k(t)}{dt^3} \right)^2 dt
\end{equation}

Since the constraints involve the position, velocity, and acceleration at both the initial ($t=0$) and terminal ($t=T_f$) instants, the optimal trajectory for each axis is analytically proven to be a fifth-order polynomial\cite{r15}:

\begin{equation} \label{eq:polynomial_def}
    p_k(t) = \sum_{i=0}^5 c_{k,i} t^i = c_{k,0} + c_{k,1}t + \dots + c_{k,5}t^5
\end{equation}
where $c_{k,i}$ are the polynomial coefficients. To determine these coefficients, we construct the boundary value problem for the full 3D state. Crucially, the terminal acceleration constraints for the longitudinal axes ($x, z$) are derived from our Stage One optimization, while the lateral axis ($y$) is regulated to align with the platform center:

\begin{equation} \label{eq:boundary_conditions}
    \begin{bmatrix} 
    \mathbf{p}(T_f) \\ \mathbf{v}(T_f) \\ \mathbf{a}(T_f) 
    \end{bmatrix} 
    = 
    \begin{bmatrix} 
    x_f & y_f & z_f \\ 
    v_{xf} & v_{yf} & v_{zf} \\ 
    a_{opt, x} & 0 & a_{opt, z} 
    \end{bmatrix}^T
\end{equation}

Here, the terminal accelerations in the longitudinal plane, $a_{opt, x}$ and $a_{opt, z}$, are the optimized values from Eq. (\ref{eq:full_optimization_problem}), ensuring the quadrotor actively achieves the ideal landing pitch. The lateral terminal acceleration is set to zero ($a_{yf}=0$) to enforce a stable, upright roll attitude relative to the platform's heading. As demonstrated in \cite{r15}, the coefficients $\mathbf{c}_k$ for all three axes are solved in closed form via a linear mapping, allowing the complete 3D trajectory to be generated in microseconds.

\subsection{Nonlinear Control of Quadcopter}
In view of the large-angle maneuvering scenarios that may exist during landing, we use the nonlinear controller proposed by \cite{r16,r17} to control the quadrotor to fly along the desired trajectory. The controller consists of a position controller and an angular rate controller. The input of the controller contains three-dimensional translation instructions and one-dimensional heading instructions. The goal of trajectory tracking is to reduce the tracking error, that is:

\begin{align}
\mathbf{e}_R = \frac{1}{2}(\mathbf{R}_d^T \mathbf{R} - \mathbf{R}^T \mathbf{R}_d)^\vee,
\end{align}
where $\mathbf{R}, \mathbf{R}_d$ represent the current attitude and the desired attitude of the quadrotor, respectively. According to \cite{r16}, in order to control the movement of the quadrotor, it can be achieved by controlling the total thrust and the desired direction $\mathbf{e}_{des}$ of the $\mathbf{e}_{3}$ axis of the body, and at the same time projecting the $\mathbf{e}_{1}$ axis of the body onto a plane perpendicular to $\mathbf{e}_{des}$ to control the heading and obtain the complete desired attitude, where the direction of $\mathbf{e}_{des}$ is determined by the direction of the desired acceleration $\mathbf{a}_{des}$. According to \cite{r17}, we can get the throttle input $\mathbf{c}_{cmd}$ and the desired angular rate $\mathbf{\omega}_{des}$ as:
\begin{align}
	\mathbf{c}_{cmd}=\mathbf{a}^T_{des}\mathbf{e}_{3}-k_h(\mathbf{v}_b^T(\mathbf{e}_{1}+\mathbf{e}_{2}))^2,
\end{align}
\begin{align}
	\boldsymbol{\omega}_{des}=\boldsymbol{\omega}_{fb}+\boldsymbol{\omega}_{ref}.
\end{align}
Here, $k_h$ is a constant, $\mathbf{v}_b$ is the speed of the quadrotor, $\mathbf{e}_{1},\mathbf{e}_{2},\mathbf{e}_{3}$ are the coordinate axes of the aircraft system, $\boldsymbol{\omega}_{fb}$ is the feedback calculated by the attitude loop, and $\boldsymbol{\omega}_{ref}$ is the feedforward given by the reference trajectory. After calculating the control amount required for the quadrotor to track the trajectory, we send the control command to the flight controller to execute the maneuver.

\begin{figure*}[t]
	\centering
	\includegraphics[width=\textwidth]{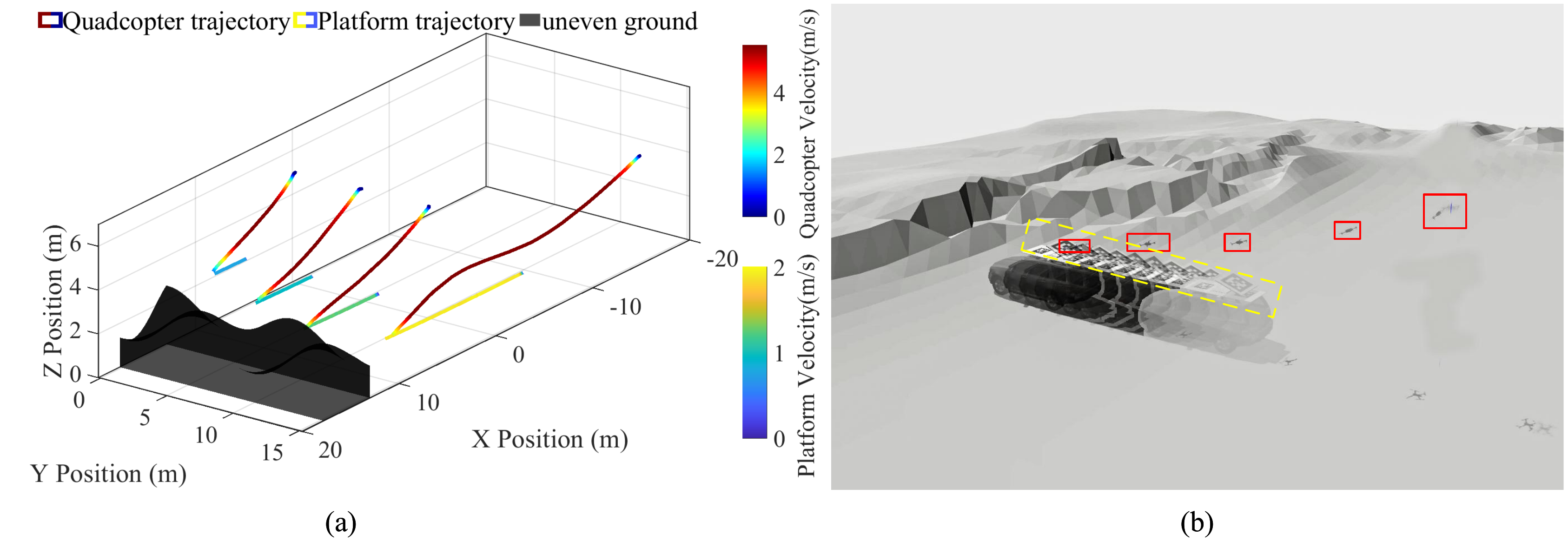}
	
	\caption{Simulation experiment results. (a) Trajectory of the quadrotor and the landing platform at different vehicle speeds. The platform moving speeds from inside to outside are $0.8\,\text{m/s}$, $1.0\,\text{m/s}$, $1.3\,\text{m/s}$, $1.5\,\text{m/s}$, and $2.0\,\text{m/s}$. (b) Afterimage of the Quadcopter (red frame) landing on the variable moving platform (yellow dotted line) at a speed of $1.0\,\text{m/s}$, and successfully landing on the platform before the AGV enters the rough ground.}
	
	\label{fig:wide_figure}
\end{figure*}

\begin{figure}[htbp]
\centering
\includegraphics[width=\columnwidth]{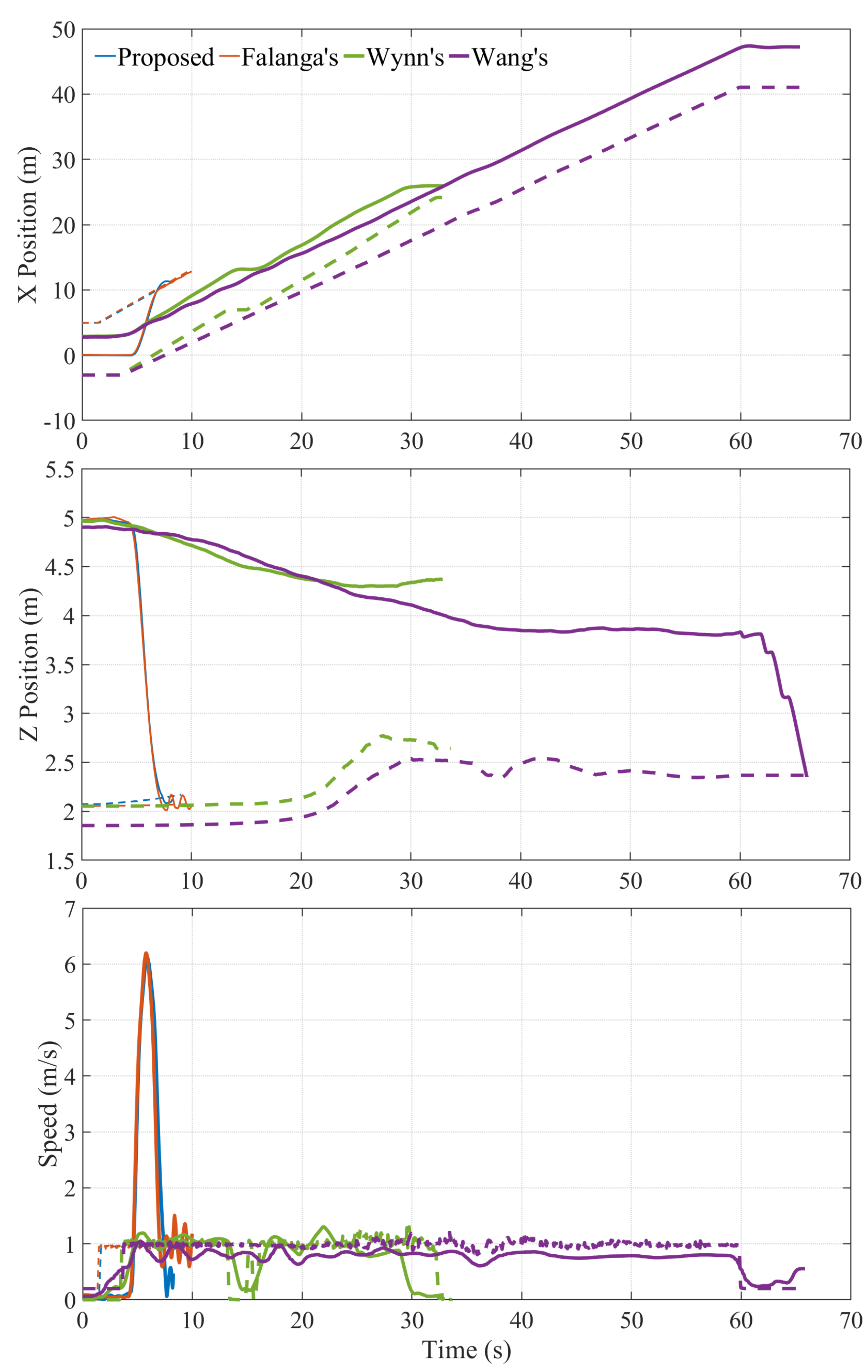}
\caption{Performance comparison of three different landing strategies at an AGV speed of $1.0\,\text{m/s}$, where solid lines represent the quadrotor trajectory and dashed lines represent the landing platform trajectory.}
\label{fig:exp_hardware}
\end{figure}


\subsection{Platform Active Control}

In the traditional visual landing system based on static AprilTags, we found that the horizontally placed AprilTag has obvious detection bottlenecks under complex lighting conditions or large-angle observation. Therefore, in our collaborative adaptive landing framework, the variable-attitude platform can not only adapt to the terminal state of the quadrotor, but also realize active visual enhancement before the quadrotor lands. Through two-way state sharing and active motion compensation between the quadrotor and the platform, the system increases the robustness of visual detection and then adjusts to the desired attitude to improve the landing success rate.

Define the detected image coordinate error as $\mathbf{e}(t)=\mathbf{z}(t)-\mathbf{z}_d$, then the error change rate can be expressed as $\dot{\mathbf{e}}(t)=\dot{\mathbf{z}}(t)$, meaning the error dynamics are determined by the target's motion in the image plane. Our goal is to adjust the angular velocity $\boldsymbol{\omega}_p$ of the landing platform so that $\mathbf{e}(t)$ converges to 0. The observation equation and observation change rate of the target in the quadcopter frame are expressed as:

\begin{align}
	\mathbf{z}(t)=\pi(\mathbf{X}_c)=\begin{bmatrix}
		f_x\frac{X_c}{Z_c}+c_x \\
		f_y\frac{Y_c}{Z_c}+c_y
	\end{bmatrix}
\end{align}
\begin{align}
	\dot{\mathbf{z}}(t)=\frac{\partial\pi}{\partial{\mathbf{X}_c}}\dot{\mathbf{X}}_c
\end{align}

Here, $\pi(\cdot)$ is the projection function of the camera, $f_x,f_y$ are the focal lengths, $c_x,c_y$ are the optical center coordinates, and $\mathbf{X}_c=[X_c,Y_c,Z_c]^T$ is the position of the target in the camera coordinate system, which can be expressed as:
\begin{align}
\mathbf{X}_c=\mathbf{R}_c^T(\mathbf{R}_p\mathbf{X}_p+\mathbf{p}_p-\mathbf{p}_b)
\end{align}
where $\mathbf{R}_c,\mathbf{R}_p$ represent the camera extrinsic rotation and platform attitude respectively, $\mathbf{X}_p$ is the position of the AprilTag in the platform system, and $\mathbf{p}_p,\mathbf{p}_b$ represent the positions of the platform and quadrotor in the world system respectively. The time derivative of $\mathbf{X}_c$ is:
\begin{align}
\dot{\mathbf{X}}_c=\mathbf{R}_c^T(\dot{\mathbf{R}}_p\mathbf{X}_p+\dot{\mathbf{p}}_p-\dot{\mathbf{p}}_b)+\dot{\mathbf{R}}_c^T(\mathbf{R}_p\mathbf{X}_p+\mathbf{p}_p-\mathbf{p}_b)
\end{align}
Substituting into (21) and rearranging yields the interaction dynamics:
\begin{align}
\dot{\mathbf{z}}(t) &=
\frac{\partial \pi}{\partial \mathbf{X}_c} \mathbf{R}_c^T (\dot{\mathbf{p}}_p - \dot{\mathbf{p}}_b)
- \frac{\partial \pi}{\partial \mathbf{X}_c} [\mathbf{X}_c]_\times \boldsymbol{\omega}_b \notag \\ 
&\quad 
- \frac{\partial \pi}{\partial \mathbf{X}_c} \mathbf{R}_c^T \mathbf{R}_p [\mathbf{X}_p]_\times \boldsymbol{\omega}_p \notag \\ 
&= \mathbf{J}_{trans} (\dot{\mathbf{p}}_p - \dot{\mathbf{p}}_b)
+ \mathbf{J}_\omega \boldsymbol{\omega}_b 
+ \mathbf{J}_p \boldsymbol{\omega}_p
\end{align}

In the above formula, $\mathbf{J}_{trans}, \mathbf{J}_\omega, \mathbf{J}_p$ reflect the dynamic coupling effect of relative translation, quadrotor rotation $\boldsymbol{\omega}_b$, and platform rotation $\boldsymbol{\omega}_p$ on the image velocity $\dot{\mathbf{z}}(t)$, respectively. Based on this, the error change rate is:
\begin{align}
\dot{\mathbf{e}}(t)=\dot{\mathbf{z}}(t)=\mathbf{J}_{trans}(\dot{\mathbf{p}}_p - \dot{\mathbf{p}}_b) + \mathbf{J}_\omega\boldsymbol{\omega}_b + \mathbf{J}_p\boldsymbol{\omega}_p
\end{align}
From this, the platform angular velocity control law is derived to compensate for these motions:
\begin{align}
\boldsymbol{\omega}_p = \mathbf{J}_p^+ \left( -K_p\mathbf{e} - \mathbf{J}_{trans}(\dot{\mathbf{p}}_p - \dot{\mathbf{p}}_b) - \mathbf{J}_\omega\boldsymbol{\omega}_b \right)
\end{align}
where $\mathbf{J}_p^+$ is the pseudo-inverse of the Jacobian matrix, and $K_p$ is the proportional control gain.

\begin{figure*}[t]
    \centering
    \begin{minipage}[t]{0.32\textwidth}
        \centering
        \includegraphics[width=\linewidth, height=4.5cm,keepaspectratio]{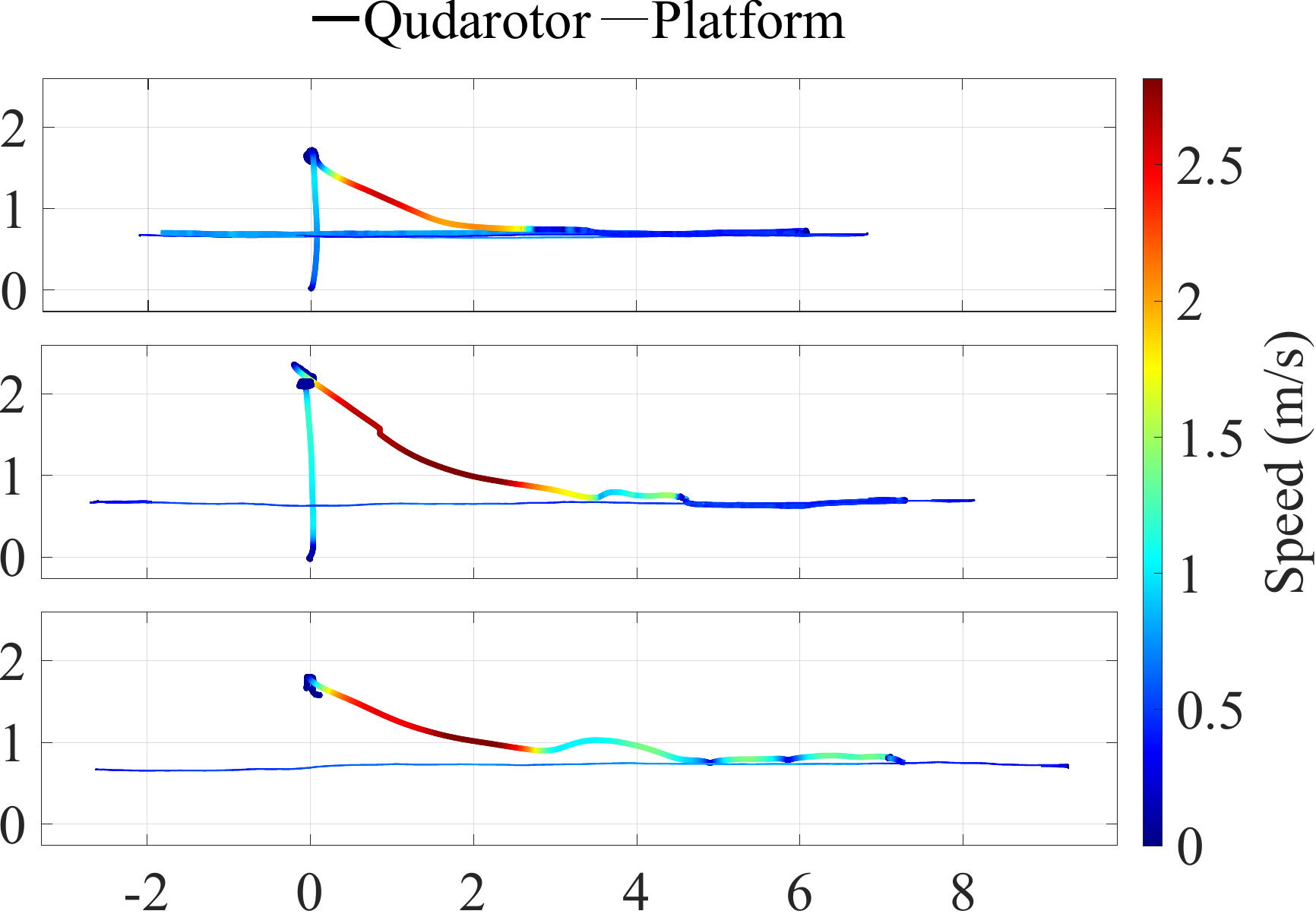} 
        \vspace{3pt} 
        \centerline{\small (a) Trajectory Results} 
        \label{fig:sub_traj}
    \end{minipage}
    \hfill
    \begin{minipage}[t]{0.32\textwidth}
        \centering
        \includegraphics[width=\linewidth, height=4.5cm,keepaspectratio]{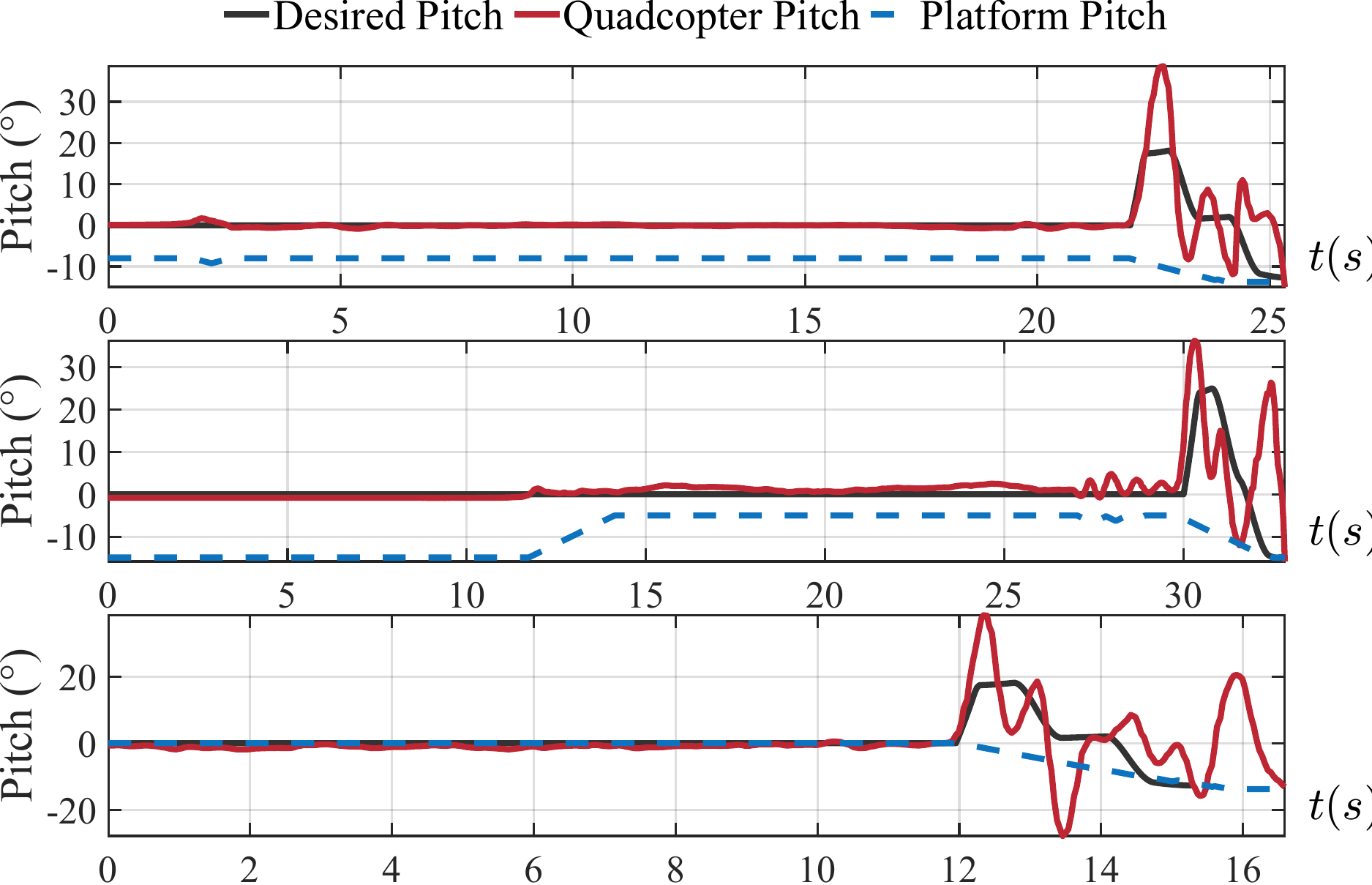}
        \vspace{3pt}
        \centerline{\small (b) Pitch Comparison}
        \label{fig:sub_pitch}
    \end{minipage}
    \hfill
    \begin{minipage}[t]{0.32\textwidth}
        \centering
        \includegraphics[width=\linewidth, height=4.5cm,keepaspectratio]{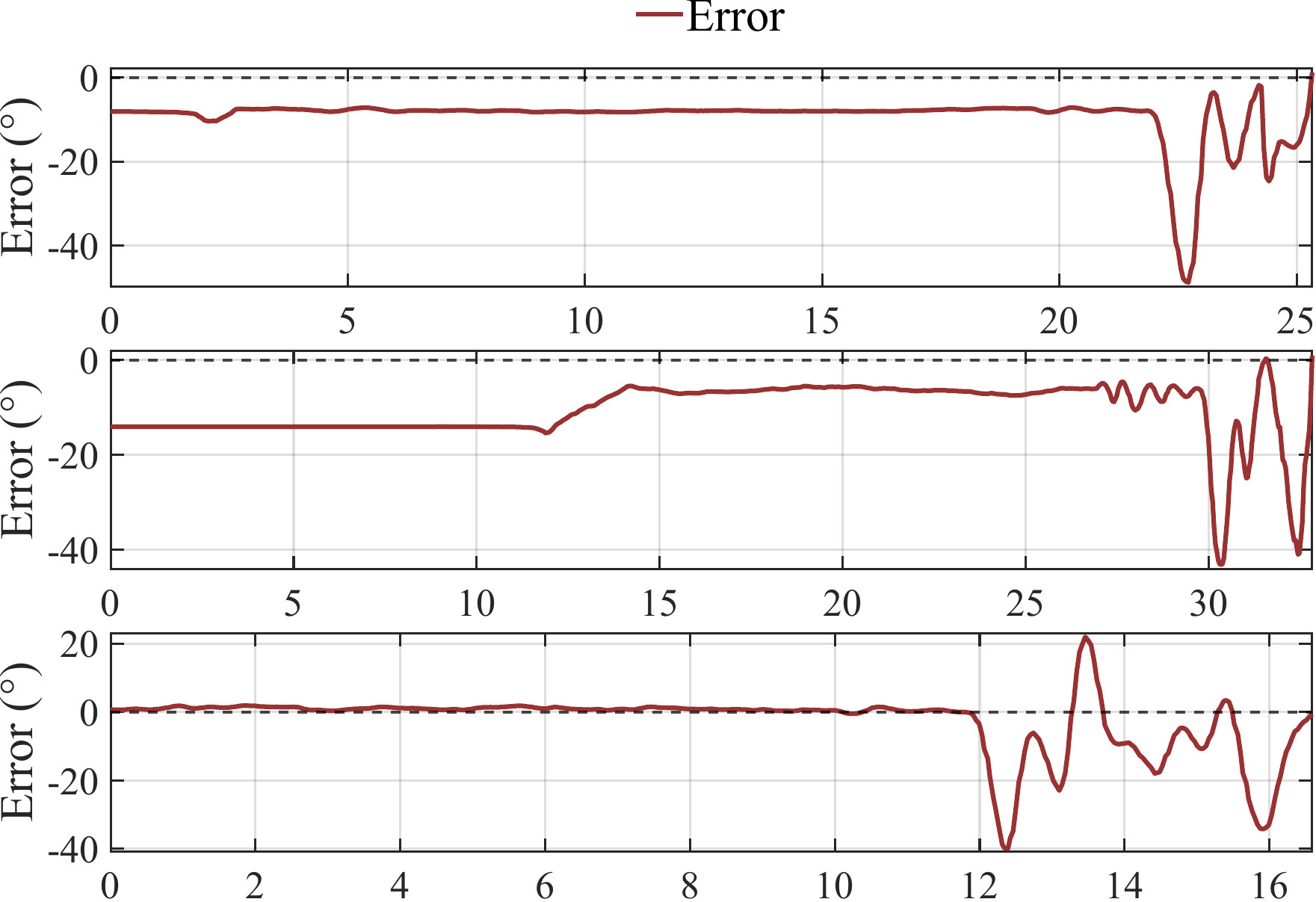}
        \vspace{3pt}
        \centerline{\small (c) Pitch Alignment Error}
        \label{fig:sub_error}
    \end{minipage}
    
    \vspace{0.2em} 
    \caption{Real experimental data results. (a) shows the trajectory, (b) compares the pitch angles, and (c) presents the relative pitch error.}
    \label{fig:wide_figure}
\end{figure*}

\begin{figure*}[t]
	\centering
	\includegraphics[width=\textwidth]{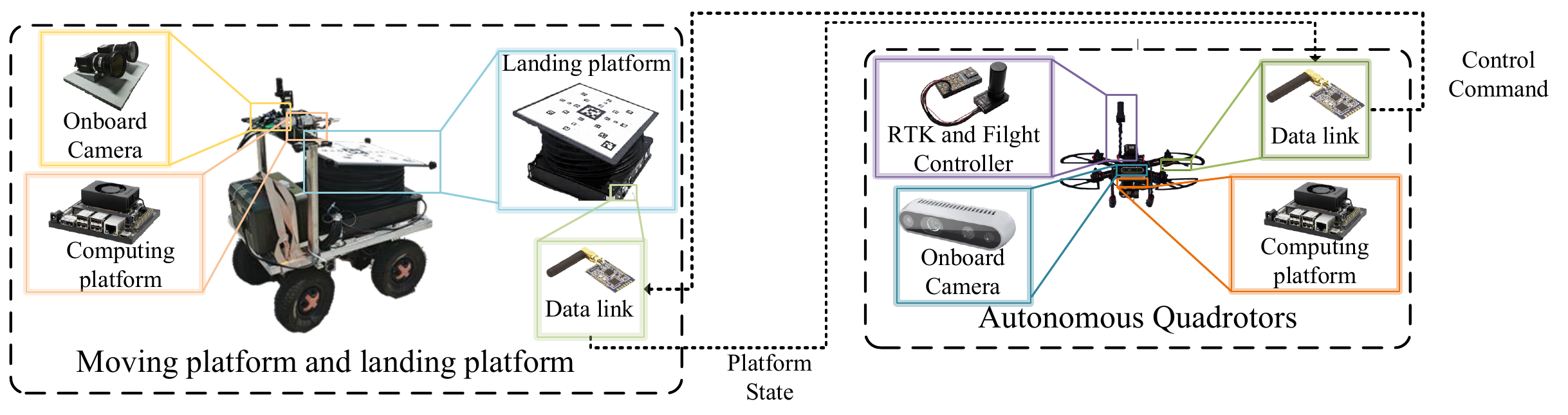}
	
	\caption{The hardware platform for cooperative autonomous landing. The left panel shows the AGV, featuring a ariable-attitude landing platform that can actively adjust its attitude to perform cooperative maneuvers. The right panel displays the quadrotor, which is outfitted with an onboard computer and sensors for autonomous flight.}
	
	\label{fig:wide_figure}
\end{figure*}

\section{EXPERIMENTAL RESULTS}

\subsection{Simulation Experiment Environment}
To fully verify the functionality and robustness of the collaborative landing framework proposed in this paper before hardware deployment, we built a Software-In-The-Loop (SITL) simulation platform based on ROS Noetic. The core of the platform is the Gazebo physics engine, which provides a controllable and repeatable high-fidelity environment for algorithmic validation.

We believe that a meaningful simulation must reproduce the constraints of real-world missions. Therefore, we deliberately designed a non-ideal operational scene featuring rugged terrain flanking a designated flat trajectory. This setup serves a specific strategic purpose: the rugged areas represent environmental ``No-Landing Zones,'' while the flat section where the mobile platform operates constitutes a strict spatial and temporal Landing Window. This design does not merely test the platform's suspension; more importantly, it imposes a hard mission constraint that forces the planner to guide the quadrotor to dock accurately only during the safe phase, validating the temporal synchronization capability of our framework.

The robot models in the simulation strive for high fidelity. The quadrotor utilizes the widely verified Iris model. The ground mobile system is modeled in detail via SDF files, featuring a mobile chassis and a variable-attitude platform controlled by independent joints. This setup accurately simulates the kinematic chain and degrees of freedom of the physical hardware. All simulation algorithms are designed to be seamlessly migrated to the physical platform, ensuring the validity and transferability of the results.

\subsection{Simulation Experiment Results}
To quantitatively evaluate the robustness of the proposed collaborative landing framework for large maneuver landing on dynamic targets, we designed two sets of progressive challenging experiments. The results are shown in Table I.

\begin{table}[t]
\centering 
\caption{The results of Cooperative Landing In simulation}
\label{table_booktabs_final}

\setlength{\tabcolsep}{10pt} 

\theadset{\bfseries} 

\begin{tabular}{c c c c}
\toprule

\thead{Plat. \\Speed\\(m/s)} & 
\thead{Des. \\Pitch\\(°)} & 
\thead{Quad. \\Pitch\\(°)} & 
\thead{Land. \\Time\\(s)} \\

\midrule
0.8  & 19.75 & 20.58 & 2.439 \\
1.0  & 19.78 & 21.11 & 3.504 \\ 
1.3  & 20.90 & 21.27 & 3.801 \\
1.5  & 21.38 & 22.70 & 3.855 \\ 
2.0  & 24.27 & 25.30 & 4.564 \\ 

\bottomrule
\end{tabular}

\vspace{0.5em} 
\footnotesize
\begin{flushleft} 
\textit{Note:} Plat. stands for Platform, Des. for Desired, Quad. for Quadrotor, and Land. for Landing.
\end{flushleft}
\vspace{-1.5em}
\end{table}

The first set is a conventional speed test: the initial relative position of the quadrotor is fixed at $5\,\text{m}$ behind the platform and $5\,\text{m}$ above the platform (i.e., relative coordinates [-5, 0, 5] meters), and its maximum speed and acceleration are limited to $5\,\text{m/s}$ and $5\,\text{m/s}^2$ respectively. The ground platform moves at a uniform linear speed at increasing speeds ($0.8\,\text{m/s}$, $1.0\,\text{m/s}$, $1.3\,\text{m/s}$, $1.5\,\text{m/s}$).

The second set is an extreme speed and long-distance test: to further explore the performance boundaries of the system, we increase the platform speed to $2.0\,\text{m/s}$ and increase the initial horizontal distance of the quadrotor to about $16\,\text{m}$(relative coordinates [-16, 0, 3] meters) to simulate an emergency pursuit scenario at a longer distance. The results are shown in Fig. 4.

To comprehensively and systematically evaluate the performance of the cooperative framework proposed in this paper, we designed rigorous comparative experiments against two methods representing different landing philosophies \cite{r8,r9}, with the experimental results summarized in Table II.

\begin{figure*}[t]
	\centering
	\includegraphics[width=\textwidth]{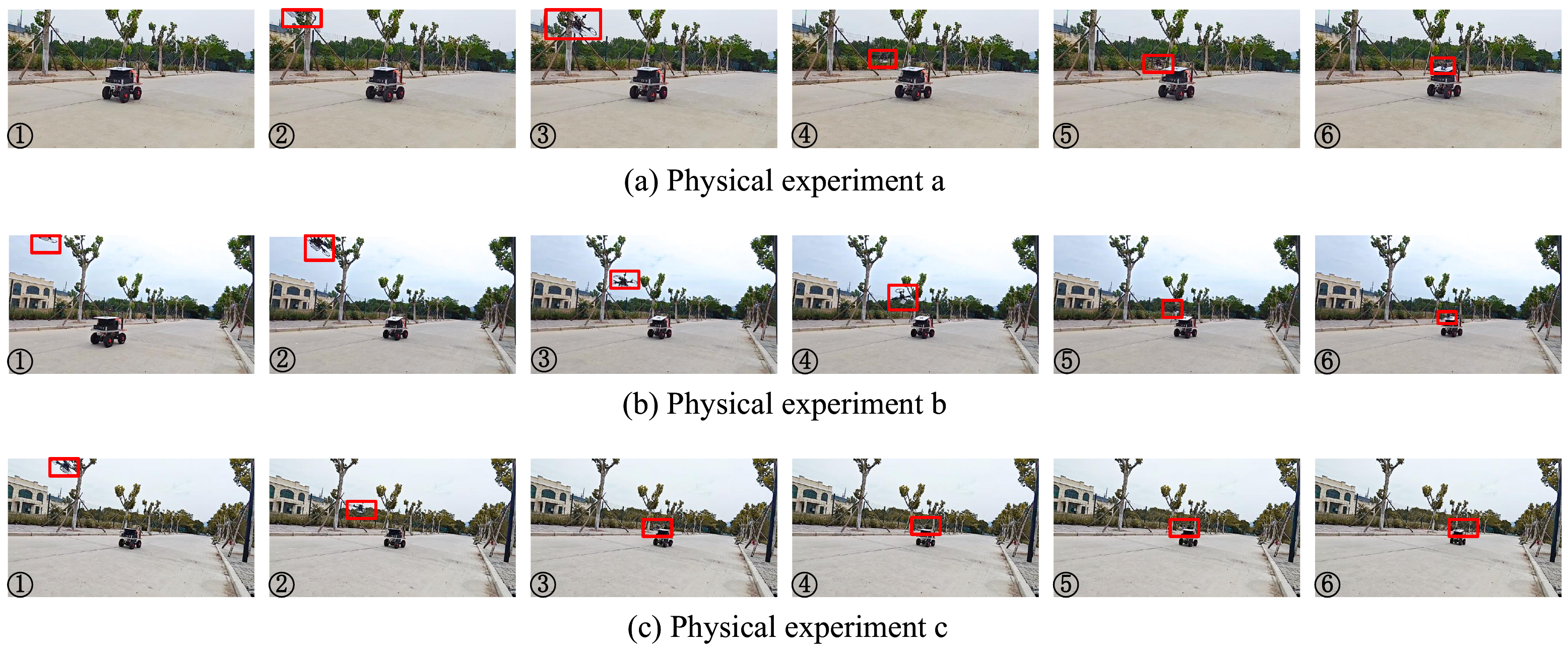}
	
	\caption{The process of the physical experiment. The quadrotor (red frame) first hovers at a specified height. After detecting the mobile landing platform, it calculates the desired pitch angle based on its own state and the state of the mobile landing platform. Then, by actively controlling the platform, the platform is adjusted to the desired attitude, realizing the landing process with two-way information sharing. In (a) Physical Experiment a, the quadrotor hovers at a height of $1.7\,\text{m}$ and the maximum speed of the mobile landing platform is $0.4\,\text{m/s}$. In (b) Physical Experiment b, the quadrotor hovers at a height of $2.3\,\text{m}$ and the maximum speed of the mobile landing platform is $0.7\,\text{m/s}$.In (c) Physical Experiment c,The platform intentionally delays its actuation. The quadrotor identifies the constraint mismatch, initiates trajectory replanning, and executes the landing only after the platform actively synchronizes its attitude}
	
	\label{fig:wide_figure}
\end{figure*}

Experimental results (Fig. 5) reveal the limitations of existing methods under physical disturbances, where high-speed platform jolting ($>1\,\text{m/s}$) induces severe tag jitter. Wynn's IBVS \cite{r8} failed as its high-gain reactive control exceeded the quadrotor's dynamic limits, leading to divergence. Similarly, Wang's method \cite{r24} achieved success but suffered from inefficiency, as its decoupled ``Hover-then-Descend'' logic necessitated prolonged hovering to satisfy strict alignment thresholds.

The comparison with Falanga's\cite{r9} unilateral planning method further reveals the true necessity of the bi-directional cooperative landing framework. This method exposed two critical flaws. First, as shown in Fig. 5, in the final stage of landing, the quadrotor exhibited significant oscillations in the Z-direction (altitude), which reflects multiple replannings upon terminal detection failure. Second, in higher-speed scenarios, to catch up with the platform within the time window, the quadrotor must adopt a very aggressive, large-pitch-angle forward flight attitude. From this extreme viewing angle, the horizontally-placed AprilTag undergoes severe perspective distortion, causing the vision recognition algorithm to fail completely, thus losing the target.

\begin{table}[t]
\centering 
\caption{Landing Time Comparison for Different Strategies}
\label{table_booktabs_final}

\setlength{\tabcolsep}{3pt} 

\resizebox{\columnwidth}{!}{%
    \begin{tabular}{c c c c c}
    \toprule
    
    \diagbox{Plat. Speed}{Method} & 
    \thead{Proposed} & 
    \thead{Falanga's} & 
    \thead{Wang's} &  
    \thead{Wynn's} \\
    
    \midrule
    
    0.8\,m/s & 2.439\,s & 2.443\,s & 25.361\,s & 32.580\,s \\
    1.0\,m/s & 3.504\,s & 3.520\,s & 67.385\,s & \XSolidBrush \\
    1.3\,m/s & 3.801\,s & 3.812\,s & \XSolidBrush & \XSolidBrush \\
    1.5\,m/s & 3.855\,s & \XSolidBrush & \XSolidBrush & \XSolidBrush \\
    2.0\,m/s & 4.564\,s & \XSolidBrush & \XSolidBrush & \XSolidBrush \\
    
    \bottomrule
    \end{tabular}%
} 

\vspace{0.2em}
\footnotesize
\begin{flushleft} 
\textit{Note:} Plat. stands for Platform, \XSolidBrush$=$mission uncompleted.
\end{flushleft}
\vspace{-1.5em}
\end{table}

In conclusion, this three-way comparative experiment leads to two key conclusions: First, for time-critical tasks, predictive trajectory planning (our method and Falanga's) is far superior in task success rate and efficiency to the reactive IBVS. Second, given an equally capable predictive planner, bi-directional cooperation is the core element for final success: through active attitude cooperation, it both solves the visual stability problem on bumpy terrain and expands the effective perception angle of the quadrotor during large maneuvers , and is the key to achieving safe, stable, and high-precision landings.

\subsection{Real-World Experiments Setups}

\begin{enumerate}[label=\arabic*), font=\bfseries, leftmargin=*, wide]
    \item {\emph{System Configuration}:} 
    Outdoor experiments were conducted using a custom quadrotor (Fig. 7, right) built with a carbon fiber frame and 3D-printed components. Equipped with 9.5-inch propellers and AIR2213 920KV motors, the system runs ROS Noetic on an NVIDIA Jetson Orin. Perception and communication rely on a D435i camera and a LoRA module, respectively. State estimation fuses RTK and PX4 IMU data via EKF, with PX4 handling inner-loop control. The 1.86 kg drone, powered by 4S batteries, achieves a thrust-to-weight ratio of 2.33.
    
    \item {\emph{Landing Platform}:} 
    We mounted a $58\,\text{cm}\times58\,\text{cm}$ variable-attitude platform on a custom heavy-duty UGV (Fig. 7, left). The platform's significant weight ($\approx 20$ kg) and 220V power requirement necessitated a torque-optimized chassis, which inherently restricts maximum speed. The system features manual remote control, LoRA-based bidirectional communication, and a $30^\circ$ upward-facing camera for quadrotor state estimation.

    \item {\emph{Relative Position Estimation}:} Inspired by \cite{r18}, we implement an EKF to robustly estimate the global states $\mathbf{x} = [\mathbf{p}_{b}^\top, \mathbf{p}_{p}^\top]^\top \in \mathbb{R}^6$.
    The prediction step propagates the state via a kinematic model driven by quadrotor odometry and platform inputs $\mathbf{u} = [\mathbf{v}_{b}^\top, \mathbf{v}_{p}^\top]^\top$.
    The update step fuses bi-directional relative position observations: the onboard detection $\mathbf{z}^b$ and the feedback $\mathbf{z}^p$.
    
    Let $\boldsymbol{\Sigma}_{\text{cam}}(z) = \text{diag}(\sigma_{pix}^2, \sigma_{pix}^2, k_z z^2)$ denote the distance-dependent sensor noise in the camera frame. The measurement covariance matrices are modeled as:
    \vspace{-0.15cm} 
    \begin{align}
        \mathbf{R}^b &= \mathbf{R}_{c}^{b} \boldsymbol{\Sigma}_{\text{cam}}(z) (\mathbf{R}_{c}^{b})^\top + \rho(\boldsymbol{\omega}) \mathbf{I}_3 \\
        \mathbf{R}^p &= \mathbf{R}_{c}^{p} \boldsymbol{\Sigma}_{\text{cam}}(z) (\mathbf{R}_{c}^{p})^\top + k_{vib} \|\mathbf{a}_{p}\| \mathbf{I}_3
    \end{align}
    where $\mathbf{R}_{c}^{b}$ and $\mathbf{R}_{c}^{p}$ represent the extrinsic rotation matrices from the respective camera frames to the body frames. 
    The term $\rho(\boldsymbol{\omega}) \propto \tanh(\|\boldsymbol{\omega}\|)$ accounts for motion blur during rapid maneuvers, while $k_{vib} \|\mathbf{a}_{p}\|$ explicitly models measurement degradation caused by vibration.
    Finally, the fusion weight $\alpha = \text{tr}(\mathbf{R}^p)/(\text{tr}(\mathbf{R}^b) + \text{tr}(\mathbf{R}^p))$ is derived to automatically prioritize the sensor with lower uncertainty.

\end{enumerate}

\subsection{Real-World Experiments Results}

To verify the effectiveness of our framework in the physical world, we conducted three sets of outdoor dynamic landing experiments using the aforementioned setup, where the quadrotor's velocity and acceleration were constrained to $3.0\,\text{m/s}$ and $3.0\,\text{m/s}^2$.  In the standard cooperative scenarios (Experiment 1 and 2), the system demonstrated high-precision synchronization. Specifically, in Experiment 1, the platform actively tilted to $13.8^\circ$ via the LoRa link, matching the quadrotor's terminal pitch ($13.75^\circ$) with a negligible error of $<0.1^\circ$.When the speed was increased to $0.7\,\text{m/s}$ in Experiment 2, robust alignment was maintained with a safety margin error of approximately $1.1^\circ$. This precise attitude matching effectively minimizes the relative tangential velocity at the moment of touchdown, thereby suppressing the risk of sideslip and rollover.

Furthermore, Experiment 3 was specifically designed to validate the system's robustness against platform actuation lag or physical constraints. As shown by the trajectory in Fig. 6(a) and the attitude comparison in Fig. 6(b), the mobile platform initially maintained a horizontal attitude (0°), simulating an extreme scenario where the platform fails to instantaneously track the optimized tilt command. During this phase, due to the significant attitude lag of the platform, the relative pose constraints required for safe geometric docking were not satisfied. Consequently, the quadrotor initiated trajectory replanning, and the landing was successfully completed only when the platform physically adjusted its pitch to match the required angle. This confirms that even if the platform cannot immediately keep up with the quadrotor's  instructions, the bi-directional cooperative mechanism ensures safety by coordinating both agents until the terminal constraints are met.

\section{CONCLUSIONS}

This work introduces a bi-directional cooperative framework for time-critical autonomous recovery. By actively coordinating the platform's attitude, our method enables a quadcopter to perform unified, large-maneuver landings within transient windows. Experiments on moving platforms validate our approach, showing significant gains in mission efficiency and control smoothness. This cooperative paradigm is key to enabling rapid, reliable quadcopter recovery in complex environments and expanding real-world multi-robot applications.


%


\ifCLASSOPTIONcaptionsoff
  \newpage
\fi



\bibliographystyle{IEEEtran}
\bibliography{bibtex/bib/IEEEexample}
%

%




\end{document}